%% file: main.tex
\documentclass[sigconf,nonacm]{acmart}
\usepackage{booktabs} 

\usepackage{CJKutf8}
\usepackage{listings}
\usepackage{tcolorbox}
\usepackage{array}
\usepackage{tabularx}
\usepackage{booktabs}
\usepackage{amsmath}
\usepackage{amssymb}
\usepackage{mathrsfs}
\usepackage{multirow}
\usepackage{colortbl}
\usepackage{threeparttable}
\usepackage{enumitem}

\definecolor{mycyan}{cmyk}{.3,0,0,0}
\definecolor{myorg}{RGB}{188,55,0}

\renewcommand{\vec}[1]{\boldsymbol{#1}} 
\DeclareMathOperator*{\argmax}{arg\,max}

\begin{document}
\title{Personalized Dialogue Generation with Diversified Traits}


\author{Yinhe Zheng}
\authornote{Please contact zhengyinhe1@163.com for the PersonalDialog dataset}
\orcid{1234-5678-9012}
\affiliation{%
  \institution{Samsung R\&D Institute of China - Beijing (SRC-B)}
  \city{Beijing}
  \state{China}
}
\email{zhengyinhe1@163.com}

\author{Guanyi Chen}
\affiliation{%
  \institution{Utrecht University}
  \city{Utrecht}
  \state{Netherlands}
}
\email{g.chen@uu.nl}

\author{Minlie Huang}
\authornote{Corresponding author}
\affiliation{%
  \institution{Tsinghua University}
  \city{Beijing}
  \country{China}}
\email{aihuang@mail.tsinghua.edu.cn}

\author{Song Liu}
\orcid{1234-5678-9012}
\affiliation{%
  \institution{Samsung R\&D Institute of China - Beijing (SRC-B)}
  \city{Beijing}
  \state{China}
}
\email{s0101.liu@samsung.com}

\author{Xuan Zhu}
\orcid{1234-5678-9012}
\affiliation{%
  \institution{Samsung R\&D Institute of China - Beijing (SRC-B)}
  \city{Beijing}
  \state{China}
}
\email{xuan.zhu@samsung.com}
\renewcommand{\shortauthors}{Y. Zheng et al.}

\begin{abstract}
Endowing a dialogue system with particular personality traits is essential to deliver more human-like conversations. However, due to the challenge of embodying personality via language expression and the lack of large-scale persona-labeled dialogue data, this research problem is still far from well-studied. In this paper, we investigate the problem of incorporating explicit personality traits in dialogue generation to deliver personalized dialogues.

To this end, {\bf firstly}, we construct \texttt{PersonalDialog}, a large-scale multi-turn dialogue dataset containing various traits from a large number of speakers. The dataset consists of 20.83M sessions and 56.25M utterances from 8.47M speakers. Each utterance is associated with a speaker who is marked with traits like \texttt{Age}, \texttt{Gender}, \texttt{Location}, \texttt{Interest Tags}, etc. Several anonymization schemes are designed to protect the privacy of each speaker. This large-scale dataset will facilitate not only the study of personalized dialogue generation, but also other researches on sociolinguistics or social science.

{\bf Secondly}, to study how personality traits can be captured and addressed in dialogue generation, we propose persona-aware dialogue generation models within the sequence to sequence learning framework. Explicit personality traits (structured by key-value pairs) are embedded using a trait fusion module. During the decoding process, two techniques, namely {\it persona-aware attention} and {\it persona-aware bias}, are devised to capture and address trait-related information. Experiments demonstrate that our model is able to address proper traits in different contexts. Case studies also show interesting results for this challenging research problem. 


\end{abstract}

%
%


\begin{CJK}{UTF8}{gbsn}
\maketitle
\input{sections/intro.tex}
\input{sections/related.tex}
\input{sections/model.tex}
\input{sections/dataset.tex}
\input{sections/experiment.tex}
\input{sections/conclusion.tex}
\bibliographystyle{ACM-Reference-Format}
\bibliography{ref}
\end{CJK}
\end{document}

%% file: sections/intro.tex
\section{Introduction} \label{lab:intro}

\newtcbox{\TraitBox}[1][red]
  {on line, arc = 2pt, outer arc = 2pt,
    colback = #1!10!white, colframe = #1!50!black,
    boxsep = 0pt, left = 1pt, right = 1pt, top = 2pt, bottom = 2pt,
    boxrule = 0pt}
    
\begin{figure}[htb!]
\small
\begin{tcolorbox}[colback = white, boxrule = 0.3mm]
A: You would rather be fashionable than comfortable. (in cold winter)

{ \parindent 1em
(真是要风度不要温度的)}

\hangafter 1
\hangindent 1em
B: Nope! I am a \TraitBox[red]{tomboy} who prefer comfortable than fashionable.

{ \parindent 1em
(才没有！我是个要温度不要风度的\TraitBox[red]{女汉子})}

\hangafter 1
\hangindent 1em
A: As your \TraitBox[green]{elder brother}, I only have one such faerie like you. You have to take care of yourself for me.

{ \parindent 1em
(\TraitBox[green]{哥哥我}就这么一个小仙女，你要替我照顾好自己)}

B: You are also in Shenzhen right?

{ \parindent 1em
(你不是也在深圳)}

\hangafter 1
\hangindent 1em
A: Yeah, I have \TraitBox[blue]{been in Shenzhen} for several years. What about you?

{ \parindent 1em
(对啊\TraitBox[blue]{在深圳}几年了，你呢) }

B: I just came to Shenzhen this year.

{ \parindent 1em
(今年刚来深圳)}

A: No wonder, we would be a couple if we live closer before.

{ \parindent 1em
(怪不得，要是近一点说不定我们都在一起了)}

\rule{\textwidth}{0.2mm}

Personality traits of A:
\begin{lstlisting}[basicstyle=\small\ttfamily]
{ "age": "24",
  `\TraitBox[green]{"gender": "Male"}`,
  "location": "Guangdong"}
\end{lstlisting}

Personality traits of B:
\begin{lstlisting}[basicstyle=\small\ttfamily]
{ "age": "23",
  `\TraitBox[red]{"gender": "Female"}`,
  `\TraitBox[blue]{"location": "Guangdong"}`}
  
\end{lstlisting}
\end{tcolorbox}
\caption{An example dialogue session (translated) in our dataset. Several personality traits are given for each speaker. Words in response are in the same color with the corresponding traits.}
\label{fig:1}
\vspace{-3mm}
\end{figure} 

Building human-like conversational systems has been a long-standing goal in artificial intelligence, where one of the major challenges is to present a consistent personality, so that the system can gain the user’s confidence and trust \citep{shum2018eliza}. Personality settings include age, gender, language, speaking style, level of knowledge, areas of expertise, or even a proper accent. The ability of exhibiting certain personality with diversified traits is essential for conversational systems to well interact with users in a more natural and coherent way \citep{Qian2017Assigning,Li2016_ACL,Kottur2017Exploring}.

Prior studies have demonstrated promising results for imitating a certain personal style in dialogue systems. Initial efforts focus on modeling characters in movie \citep{Danescu2011_Cornell,Banchs2012Movie}. Further developments propose to use a speaker embedding vector in neural models to capture the \textit{implicit} speaking style of an individual speaker  \citep{Li2016_ACL,Kottur2017Exploring,zhang2017neural,ouchi2016addressee,zhang2017addressee}, or the style of a group of speakers \citep{Wang2017Group}. Other approaches also attempt to endow dialogue models with personae which are described by natural language sentences \citep{P18-1205,mazare2018training}.

Recent studies on personalized neural conversational models can be broadly classified into two types: one is \textit{implicit personalization} and the other is \textit{explicit personalization}. In \textit{implicit personalization models} \cite{Li2016_ACL,Kottur2017Exploring,zhang2017neural}, each speaker is represented by a user vector, and the vector is then fed into the decoder to capture the speaking style of the speaker implicitly. In spite of the simplicity and success of this technique, it is unclear how personality is captured and how it can be interpreted because all the information regarding to a user is encoded in a real-valued vector. Moreover, these methods also suffer from the data sparsity issue: each dialogue should be tagged with a speaker identifier and there should be a sufficient amount of dialogues from each speaker to train a reliable user-specific model. In {\it explicit personalization models}, the generated responses are conditioned either on a given personal profile \cite{Qian2017Assigning}, or on a text-described persona \cite{P18-1205}. In these models, personality is presented specifically via key value pairs or natural language descriptions about age, gender, hobbies, etc. However, these methods are limited to either manually-labeled data or crowdsourced dialogues, thereby not scalable to large-scale dialogue datasets.

It is a matter of fact that the persona of a speaker can be viewed as a composite of diversified personality traits. During conversations, people may reveal their personality traits, consciously or unconsciously. For example, for the dialogue shown in Figure \ref{fig:1}, speaker $B$ uses the word ``tomboy'' in response to speaker $A$'s comment. It can be inferred that speaker $B$ is a female. Similarly, based on the second and the third turns of this session, we can easily infer that both speaker $A$ and $B$ are living in Shenzhen (a city in Guangdong province, China). As exemplified, {a personalized conversational agent should be equipped with diversified traits and be able to decide which personality traits to express in different contexts}. 

To address above issues, we propose a novel task and construct a large-scale dialogue corpus to study personalized dialogue generation. The task and corpus are unique in several aspects:

\begin{itemize}[leftmargin=1em]
\item 
\textbf{First}, the persona of each speaker in the corpus is presented by a number of personality traits, which are given explicitly in key-value pairs (as exemplified in Figure \ref{fig:1}). Unlike implicit personalization models, such structured personae are more explicit, straight-forward, and interpretable. Moreover, since speakers with the same trait value (e.g., all females) can share their trait representations, the dialogue data across speakers can be shared to train a generation model, thereby the data sparsity issue is alleviated.

\item
\textbf{Second}, although the persona is represented explicitly, the use of such persona information can be captured implicitly by data-driven methods that are scalable to large-scale corpora. This differs from prior explicit personalization models~\cite{Qian2017Assigning} which require that the given persona values must appear in a generated response and demand for manually-labeled data.

\item
\textbf{Third}, it is interesting to study how personality traits are expressed in dialogues and revealed via language expressions. In fact, the expression of persona via language is usually subtle and implicit\citep{bamman2014gender}. 
For instance, a female speaker may not necessarily use the word ``female'' directly in every utterance she responds with, instead, she may consciously or unconsciously use related words that can reveal her gender in particular contexts. Therefore, it is worthy to build a personalized conversational system with the ability to exhibit specific traits in different contexts.
\end{itemize}

In this paper, we employ the sequence to sequence learning framework \citep{sutskever2014sequence,vinyals2015neural} and devise a trait fusion module to capture the persona of each speaker in the response generation process. Specifically, each trait of a speaker is encoded as an embedding vector and different traits are merged to produce an integrated persona representation. Two approaches are devised to leverage the persona representation in the generation process: the first approach introduces an persona aware attention mechanism where the persona representation is used to generate the attention weights to obtain the context vector at each decoding position, and the second approach applies an persona-aware bias to estimate the word generation distribution. Automatic and manual evaluation indicate that our proposed models can incorporate proper, diversified traits when generating responses in different contexts.

Since there is no existing corpus to facilitate the aforementioned research task, we construct a large-scale dialogue dataset which contains various personality traits for a large number of speakers. Our dataset is collected from Weibo and contains about 20.83 million dialogue sessions (in Chinese) from about 8.47 million speakers. These dialogues cover a wide range of topics about our daily lives and consist of more than 3.43 million multi-turn sessions (each containing no less than 4 utterances). Various of personality traits are collected for each speaker and three of which are approached and evaluated in our model, namely \texttt{Gender}, \texttt{Age}, and \texttt{Location}. The proposed dataset will be useful not only for the study of dialogue systems, but also for other research topics such as pragmatics or sociolinguistics.

Our main contributions can be summarized as follows:
\begin{enumerate}
  \item We propose a new task to incorporate explicit personality traits into conversation generation. This task aims to study how explicit personality traits can be used to train a personalized dialogue model with large-scale, real social conversations.
 
  \item We construct a large-scale dialogue dataset that contains various traits of each speaker
  (such as \texttt{Age}, \texttt{Gender}, \texttt{Location}, \texttt{Interest Tags} etc.).
  To the best of our knowledge, this is the first dialogue corpus that contains real social conversations and diversified personality traits for each speaker.
  The proposed dataset will facilitate not only the study of personalized dialogue generation, but also other researches such as sociolinguistics.

  \item We propose persona-aware models which apply a trait fusion module in the encoder-decoder framework to capture and address personality traits in dialogue generation. We devise a persona-aware attention mechanism and persona-aware bias to incorporate the persona information in the decoding process. Experiments demonstrate that our model is able to address proper traits in different contexts.
  
 \end{enumerate}

%% file: sections/related.tex
\section{Related Work}
It has been demonstrated that personality is vital for building a human-like dialogue system  \citep{hernault2008generating,shum2018eliza} which can exhibit a consistent persona. Personality settings such as age, gender, level of knowledge, and personal interests can be implicitly or explicitly expressed during the conversations \citep{shum2018eliza}. In order to deliver more intelligent conversations, it is thus necessary to model these personality traits properly in a personalized conversational system.

There have been various prior studies for personalized dialogue generation. Traditional models are proposed to build personalized dialogue systems by modeling the ``\emph{Big Five}'' \citep{goldberg1993structure}. This concept has been well defined in psychology \citep{norman1963toward}, and is proved to be a stable personality evaluation metric \citep{cobb2012stability}. Some personalized dialogues systems were built upon the basis of ``\emph{Big Five}'', such as \textsc{Personage} \citep{mairesse2007personage,mairesse2008personality} and the work of \citet{gill2012perceptions}. However, such personality metric is extremely implicit and subtle in language expression, and thus challenging to be captured in dialogue generation \citep{Qian2017Assigning}. Moreover, the dialogue data with ``\emph{Big Five}'' annotation are extremely complex and expensive to collect. Therefore, ``\emph{Big Five}'' is not suitable for building large-scale personalized dialogue systems, particularly with data-driven neural models.

Recently, the availability of large-scale dialogue corpora has significantly advanced the research of data-driven personalized dialogue models \citep{Kottur2017Exploring}. Some early studies focused on modeling characters in movie dialogues \citep{Danescu2011_Cornell,Banchs2012Movie}, in which the presented ``Character Style'' usually depends on the scenes and plots of each movie. Further development of personalized dialogue generation models is inspired by the successful application of social media data \citep{ritter2011data,Serban2015_data_survey} and the sequence to sequence learning framework \citep{sutskever2014sequence,vinyals2015neural,sordoni2015hierarchical,shang2015neural,serban2016building}. Specifically, \citet{Li2016_ACL} represented each speaker with a persona vector and fed the vector to the decoder at each decoding step. The persona embedding is supposed to capture speaker-specificed styles. \citet{Kottur2017Exploring} extended this idea to multi-turn dialogues. In these models, the persona is implicitly represented by a single real-valued vector, which lacks interpretability.

In spite of the success of user embedding in above models, training these models requires abundant dialogue data from each speaker. When there are no such data available, it is unlikely to train a reliable model. A possible attempt to deal with this issue is to train personalized models with the gender attribute \citep{Wang2017Group}. This approach helps to alleviate the data sparsity issue since the dialogue data within a group of same gendered speakers can be shared.

Note that personality traits in these embedding-based approaches are modeled implicitly. An initial attempt to incorporate explicitly represented persona is proposed by \citet{Qian2017Assigning}, in which a chatbot is endowed with a persona defined by a key-value table. A pair of forward and backward decoders are used to generate a response starting from a selected profile value (e.g., \textit{female}), which ensures that a selected value must appear in a generated response. This approach requires manually-labeled data, and it may not be scalable to the large dialogue dataset as proposed in this paper. It is also expensive to collect large-scale dialogue data via crowdsourcing services \cite{P18-1205}.


The construction of large-scale dialogue datasets is another important topic for recent research on dialogue systems. \citet{Serban2015_data_survey} presents a comprehensive summary of available dialogue datasets that can be used to construct a data-driven dialogue model. However, most of the existing corpora are not suitable for the study of personalized dialogue generation. Initial efforts collect dialogues from movie scripts \citep{Danescu2011_Cornell,Walker2012_corpus}, with the annotations of \emph{Character Styles}. \citet{P18-1205} crowd-sourced a dataset by asking randomly paired crowd workers to chat based on some given personae, however, this dataset is limited in its small size. The dataset introduced by \citet{Qian2017Assigning} has a similar personality trait format (i.e., key-value pairs) with ours, and consists of manually-labeled data but only covers a small amount of patterns for a few traits, which is thus not scalable to large datasets. The dataset proposed by \citet{Joshi2017Personalization} is constructed using limited templates and thus not suitable for dialogue generation tasks. We believe the dataset presented in this paper will offer new possibilities for studying personalized dialogue models with large-scale, real social conversation data.

%% file: sections/model.tex
\section{Model} \label{sec:model}
In order to capture diversified personality traits in the response generation process, we equip the general sequence to sequence model with a personality trait fusion module, which produces a persona representation $\vec{v}_p$ that can be incorporated into the decoder. In this study, two methods are proposed to utilize $\vec{v}_p$ in the decoding process, one is a persona-aware attention mechanism, and the other is a persona-aware bias. We will present the details in this section. 

\begin{figure*}[t]
    \centering
    \includegraphics[width=400px]{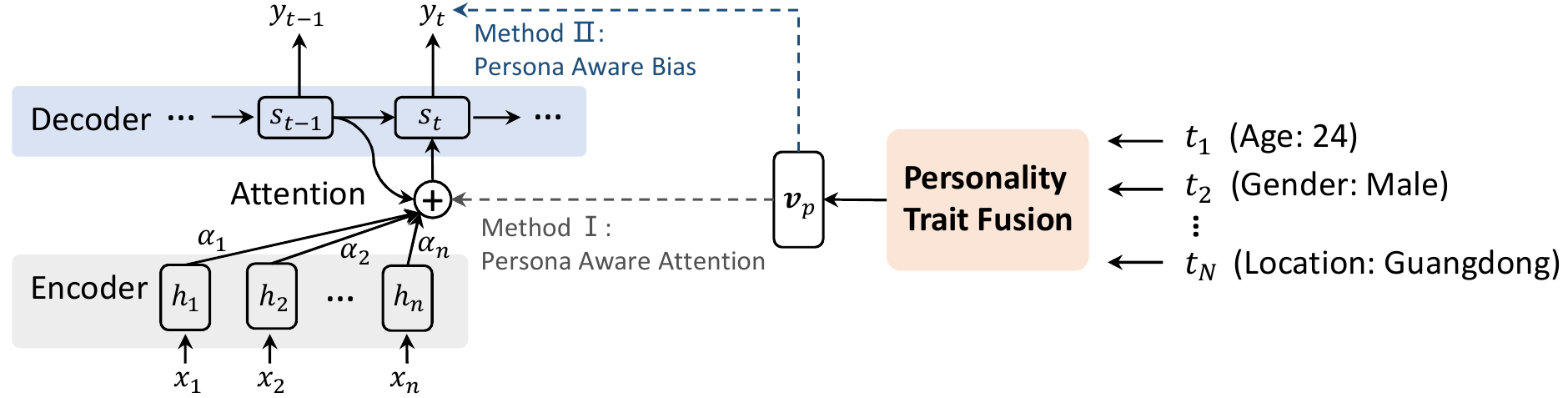}
    \caption{Overview of personalized dialogue generation model. To obtain the persona representation $\vec{v}_p$, different traits are integrated by the personality trait fusion component.  $\vec{v}_p$ is then used to generate persona-aware attention weights for computing the context vector, or to produce a persona-aware bias for computing the generation distribution.}
    \label{fig:personal_attention}
\vspace{-2mm}
\end{figure*}

\subsection{Task Definition and Overview}
Our task can be formulated as follows: Given a post $X = x_1,x_2,...,x_n$ and a set of traits  $T = \{t_1, t_2, ..., t_N\}$ for the responder, the system should generate a response $Y = y_1,y_2,..., y_m$ that embodies the personality traits in $T$:
\begin{equation}
    Y^* = \argmax_Y P(Y|X,T)
\end{equation}
Where $x_i$, $(i=1,2,...,n)$ and $y_i$, $(i=1,2,...,m)$ are words. Note that each trait $t_i \in T$ is given as a key-value pair $t_i = \langle k_i, v_i \rangle$, and the exact values of $v_i (i=1,2,...,N)$ are not required to appear in $Y$. Moreover, although the personality traits of the speaker who makes the post are also provided in our dataset, they are not modeled in our task. We leave this as future works.

An overview of our personalized dialogue generation model is shown in Figure \ref{fig:personal_attention}. Given a trait set $T$, a personality trait fusion module is used to merge the traits in $T$ into a persona representation $\vec{v}_p \in \mathbb{R}^{d_p}$. Three approaches are proposed to fuse the personality traits. A sequence encoder is used to encode the post into a series of real-valued vectors $(\vec{h}_1, \vec{h}_2, ..., \vec{h}_n)$, where $\vec{h}_i \in \mathbb{R}^{d_s}$. Two methods are proposed to incorporate $\vec{v}_p$ into the decoding process: the first method introduces a persona-aware attention, namely, using $\vec{v}_p$ to generate the attention weights at each decoding position such that the context vector computed at each position is conditioned on $\vec{v}_p$; the second method applies a persona-aware bias directly in estimating the generation distribution.

\subsection{Sequence to Sequence Framework}
The backbone of our model is the sequence to sequence (Seq2Seq) learning framework \citep{sutskever2014sequence,vinyals2015neural}, which is commonly used in language generation tasks such as machine translation and dialogue generation. A typical Seq2Seq model usually consists of two components: an encoder and a decoder. For dialogue generation tasks, the encoder takes the post $X = x_1,x_2,...,x_n$ as input and encodes $X$ into a sequence of vectors $(\vec{h}_1, \vec{h}_2, ..., \vec{h}_n)$, $\vec{h}_i \in \mathbb{R}^{d_s}$. The decoder will sample a word from a generation distribution over the vocabulary at each decoding step. The generation distribution is conditioned on the preceding state of the decoder, the previously generated word, and the context vector which is computed with an attention mechanism.

In this study, we use the attention mechanism proposed by \citet{bahdanau2014neural}, which produces a context representation $\vec{c}_t \in \mathbb{R}^{d_s}$ at each decoding step $t$ by attending to the encoder's outputs $\vec{h}_1, \vec{h}_2, ..., \vec{h}_n$, at the same time conditioned on the preceding state of the decoder $\vec{s}_{t-1} \in \mathbb{R}^{d_s}$. Formally, we have:
\begin{equation}\label{eq:score}
  \begin{split}
    \vec{c}_t &= \sum_{i=1}^{n} \alpha_i \vec{h}_i,\\
    \alpha_i  &= \frac{\exp{(e_i)}}{\sum_{j=1}^n \exp{(e_j)}}\\
    \quad e_i &= \mbox{MLP}(\vec{s}_{t-1}, \vec{h}_i)\\
              &= V^T \cdot \mbox{tanh} (W_\alpha^1\vec{s}_{t-1} + W_\alpha^2\vec{h}_i)
  \end{split}
\end{equation}
where $V \in \mathbb{R}^{d_s}$, $W_\alpha^1 \in \mathbb{R}^{d_s \times d_s}$ and $W_\alpha^2 \in \mathbb{R}^{d_s \times d_s}$ are parameters for the attention mechanism.

In general Seq2Seq models, the output probability $y_t$ at step $t$ of the decoder is produced by a softmax function:
\begin{equation} \label{eq:softmax}
  \begin{split}
    y_t               &= \mbox{softmax}(W_o^1\vec{s}_{t} + b_{out}), \\
    \quad \vec{s}_{t} &= \mbox{RNN}(\vec{s}_{t-1}, \vec{c}_t, w_{t-1}).
  \end{split}
\end{equation}
where $w_{t-1}$ is the word vector of the decoded word from previous time step. $W_o^1 \in \mathbb{R}^{|V|\times d_s} $ and $b_{out} \in \mathbb{R}^{|V|}$ are parameters for the decoder ($|V|$ is the vocabulary size).

In this study, the encoder we use is a two-layer bi-directional RNN with gated recurrent units (GRU) \citep{cho2014learning}, and the decoder is also a two-layer GRU.

\subsection{Personality Trait Fusion}\label{sec:trait_fusion}

In our personalized dialogue model, we first compute an integrated persona representation $\vec{v}_p$ and then use $\vec{v}_p$ to affect the decoding process. The construction of $\vec{v}_p$ starts with mapping each traits $t_i$ in $T$ to an embedding representation $\vec{v}_{t_i}$ using its corresponding trait encoder. 
Note that traits considered in this study (i.e., \texttt{Age}, \texttt{Gender} and \texttt{Location}) are all single-valued, i.e., each trait only has one unique value for each speaker. Therefore these trait encoders can be implemented using look-up tables.
Actually, other categories of traits can also be modeled if a proper encoder is provided. For instance, an LSTM encoder can be applied to represent a one-sentence self-description of a speaker.



After encoding all the traits in $T$ into a set of trait representations $\{\vec{v}_{t_1}, \vec{v}_{t_2}, ..., \vec{v}_{t_N}\}$, we can merge them using a \emph{personality trait fusion function} to obtain the persona representation $\vec{v}_p$. In this paper, three different fusion methods are investigated.
    
\subsubsection{\textbf{Traits Attention}}\label{sec:trait_att}
Merge all the trait representations in $T$ based on an attention mechanism. Specifically, given the hidden state from the previous decoding step $\vec{s}_{t-1}$, an attention weight $\alpha_i'$ is computed for each trait. Then, $\vec{v}_p$ is obtained as a weighted sum of all the trait representations:

\begin{equation}
  \begin{split}
    \vec{v}_p & = \sum_{i=1}^{N} \alpha_{i}' \vec{v}_{t_i} \\
    \alpha_{i}' &= \frac{\exp{(e_i')}}{\sum_j^N \exp{(e_j')}} \\
     \quad e_i' &= \mbox{MLP}(\vec{s}_{t-1}, \vec{v}_{t_i}) \\
      &= \overline{V}^T \cdot \mbox{tanh} (\overline{W_\alpha^1}\vec{s}_{t-1} + \overline{W_\alpha^2}\vec{v}_{t_i})\\
  \end{split}
\end{equation}
where $\overline{V} \in \mathbb{R}^{d_s}$, $\overline{W_\alpha^1} \in \mathbb{R}^{d_s \times d_s}$ and $\overline{W_\alpha^2} \in \mathbb{R}^{d_s \times d_p}$ are parameters for the trait attention mechanism. The calculated weight $\alpha_{i}'$ indicates how much the current context favors trait $t_i$.
The trait attention mechanism here allows us to make proper combination of personality traits with respect to the contexts.
    
\subsubsection{\textbf{Traits Average}}\label{sec:trait_avg} 
Average all the trait representations in $T$:
\begin{equation}
    \vec{v}_p = \frac{1}{N} \sum_{i=1}^{N} \vec{v}_{t_i}
\end{equation}
This is a special case of \emph{\textbf{Traits Attention}}, where the traits in $T$ are weighted equally.

\subsubsection{\textbf{Traits Concatenation}}\label{sec:trait_concat} 
Concatenate all the trait representations  in $T$ to produce $\vec{v}_p$. Note that in this case the length of $\vec{v}_p$, i.e., $d_p$ should be divisible by $N$ and the length of each trait representation vector $\vec{v}_{t_i}$ should be $d_p/N$.

\subsection{Decoding with Persona Representation}\label{sec:persona_decoder}
In order to incorporate the persona representation $\vec{v}_p$ in our decoder, we develop the following methods:

\subsubsection{\textbf{Persona-Aware Attention} (PAA)}\label{sec:paa}: 
The first method extends the computation of attention weights (Equation \ref{eq:score}) used in the decoder. The attention weight is now dependent on not only the decoder's state, but also the persona representation $\vec{v}_p$, namely,
\begin{equation}
  \begin{split}
    e_i &= \mbox{MLP}(\vec{s}_{t-1}, \vec{h}_i, \vec{v}_p) \\
    & = V\cdot \mbox{tanh} (W_\alpha^1\vec{s}_{t-1} + W_\alpha^2\vec{h}_i + W_\alpha^3\vec{v}_p)
  \end{split}
\end{equation}
where $V \in \mathbb{R}^{d_s}$, $W_\alpha^1 \in \mathbb{R}^{d_s \times d_s}$, $W_\alpha^2 \in \mathbb{R}^{d_s \times d_s}$, and $W_\alpha^3 \in \mathbb{R}^{d_s \times d_p}$ are learnable parameters. The score $e_i$ is the input to the softmax function for computing the attention weight. This approach can help our decoder to attend to different contexts based on the persona representation, which is termed persona-aware attention mechanism.

\subsubsection{\textbf{Persona-Aware Bias} (PAB)}\label{sec:pab}: 
The second method tries to incorporate $\vec{v}_p$ in the output layer of the decoder. Specifically, we extend Equation \ref{eq:softmax} to include a persona bias to obtain the generation distribution. A gate is devised to balance the original term and the persona bias term, as follows:
\begin{equation}
  \begin{split}
    y_t &= \mbox{softmax}(a_t \cdot W_o^1\vec{s}_{t} + (1-a_t) \cdot W_o^2\vec{v}_p + b_{out})\\
    a_t &= \sigma(V_o^T \cdot \vec{s}_t)
  \end{split}
\end{equation}
where $W_o^1 \in \mathbb{R}^{|V|\times d_s} $, $W_o^2 \in \mathbb{R}^{|V| \times d_p}$, $V_o \in \mathbb{R}^{d_s}$ and $b_{out} \in \mathbb{R}^{|V|}$ are learnable parameters. Note that although the bias brought by $\vec{v}_p$ seems to be context independent (i.e., it may select words independently at each decoding step), the computed scalar variable $a_t \in [0,1]$ works as a gate to control how much persona related features should be incorporated at each time step $t$. It can decide whether to use trait related word or semantic related word, and thus makes the response generation process more consistent.

As can be seen, the persona-aware bias is assumed to be more direct in influencing the generation distribution, which is verified by experiment results shown in \S\ref{sec:experiments}: PAB works generally better than PAA. Similar model structures have also been used in the work of \citet{jaech2017lowrank_rnn} and \citet{zhou2017emotional}, and promising results have achieved.

%% file: sections/dataset.tex
\section{\texttt{PersonalDialog} Dataset} \label{data_construction}
The dialogue dataset that we construct for the proposed task, named as \texttt{PersonalDialog}, involves a large amount of speakers with a wide variety of personality traits. The data in \texttt{PersonalDialog} are collected from Weibo\footnote{\url{www.weibo.com}}, one of the largest Chinese social media. In fact, according to the theories in sociolinguistics, people tend to perform specific personae when they use language to socialize \citep{goffman1959presentation,shulman2016presentation}. Therefore, social media becomes an ideal source to collect large-scale dialogues with diversified personality traits. The features and statistics of \texttt{PersonalDialog} are detailed in this section together with a brief introduction to the data collection process.

\subsection{Features and Statistics} \label{sec:features}
Dialogues in our dataset are composed of Weibo posts and their comments. Specifically, when a user post a Weibo message, other users may comment on it, which may receive further comments. It forms a tree structure which is rooted at the original Weibo post. We regard an original post and one branch of its comments as a dialogue session. These dialogues are collected along with the publicly available personality traits of each speaker. Some attractive features of \texttt{PersonalDialog} are presented in this section.

\subsubsection{Personality Traits}
The most important and appealing property of \texttt{PersonalDialog} is the personality traits collected for each speaker, which are provided by speakers themselves on Weibo. Various interesting tasks can be investigated with the help of these information, such as personalized dialogue generation, text style transfer or text-based personality analysis.

\begin{table}[htbp]
\centering
 \caption{Statistics of personality traits in \texttt{PersonalDialog}.}
 \begin{tabular}{ll}
  \toprule
  Total number of speakers             & 8.47M \\
  Total number of interest tags        & 39.6K  \\
  Number of interest tags per speaker  & 2.187  \\
  Average speaker age                  & 25.23  \\
  Average length for self descriptions & 10.09  \\
  \bottomrule
 \end{tabular}
\label{tab:traits}
\vspace{-1.5mm}
\end{table}

\begin{figure*}[htbp]
  \centering
  \begin{tabular}{ccc}
      \includegraphics[scale=0.4]{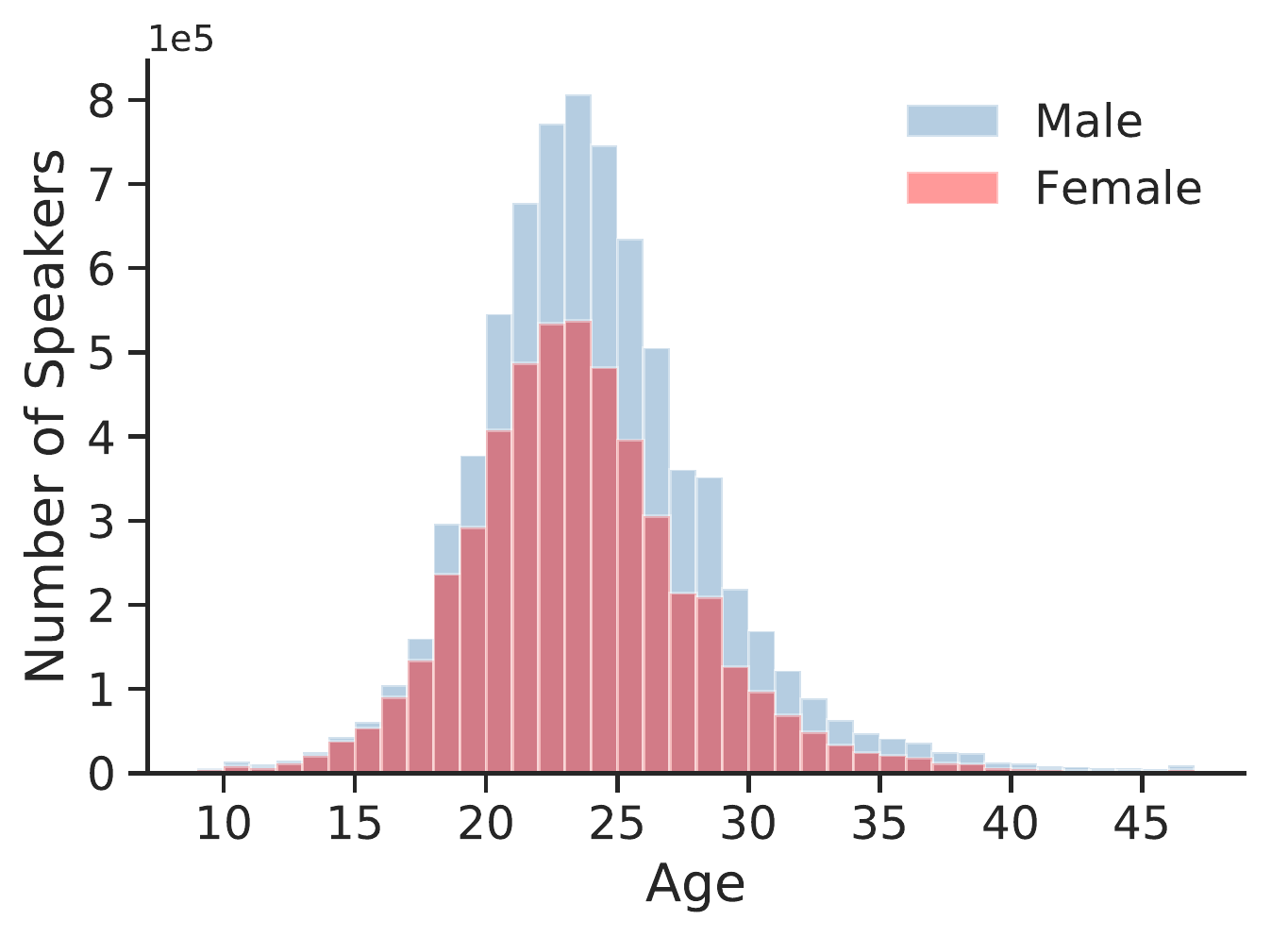} &
      \includegraphics[scale=0.35]{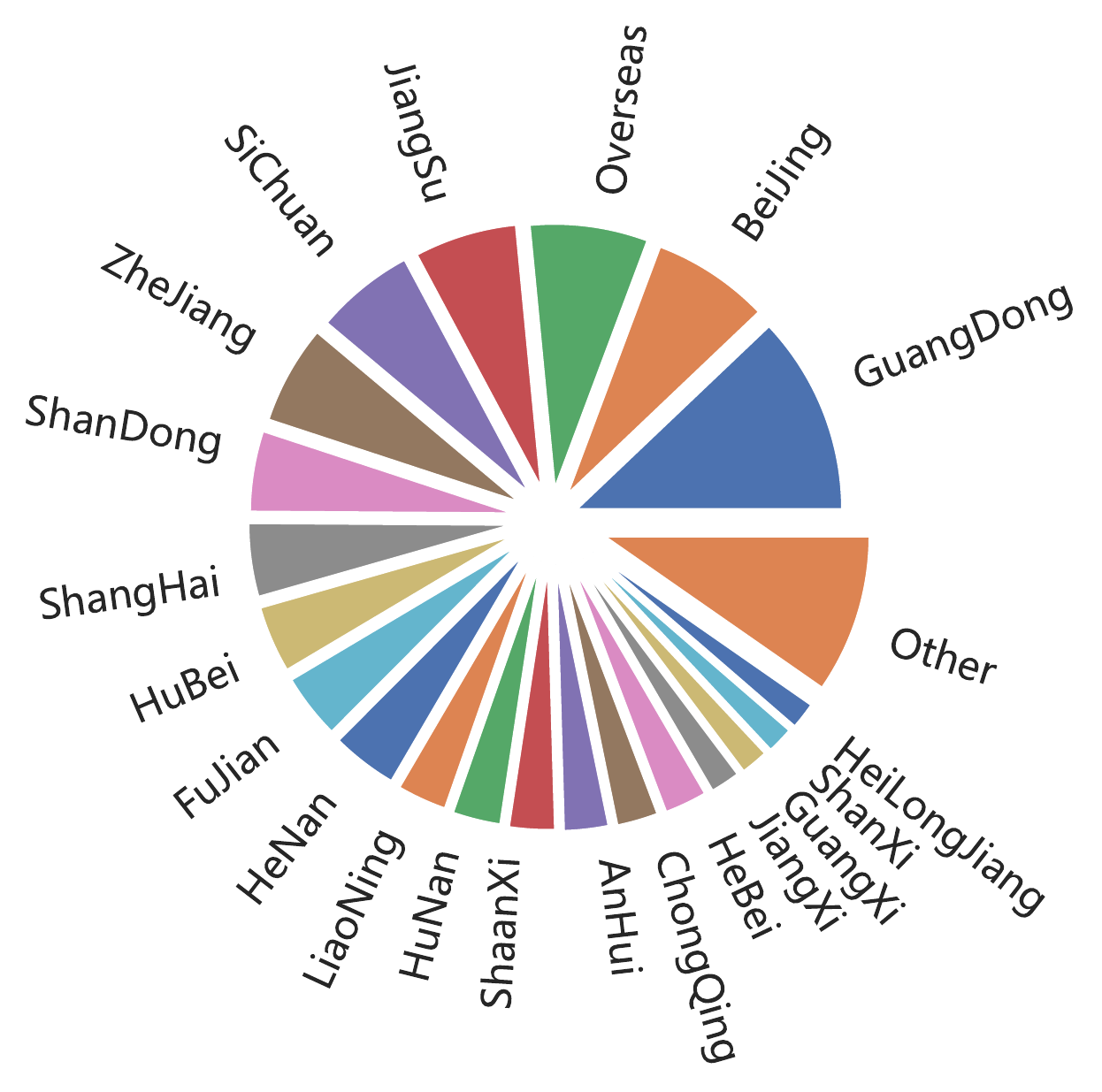} &
      \includegraphics[width=150px]{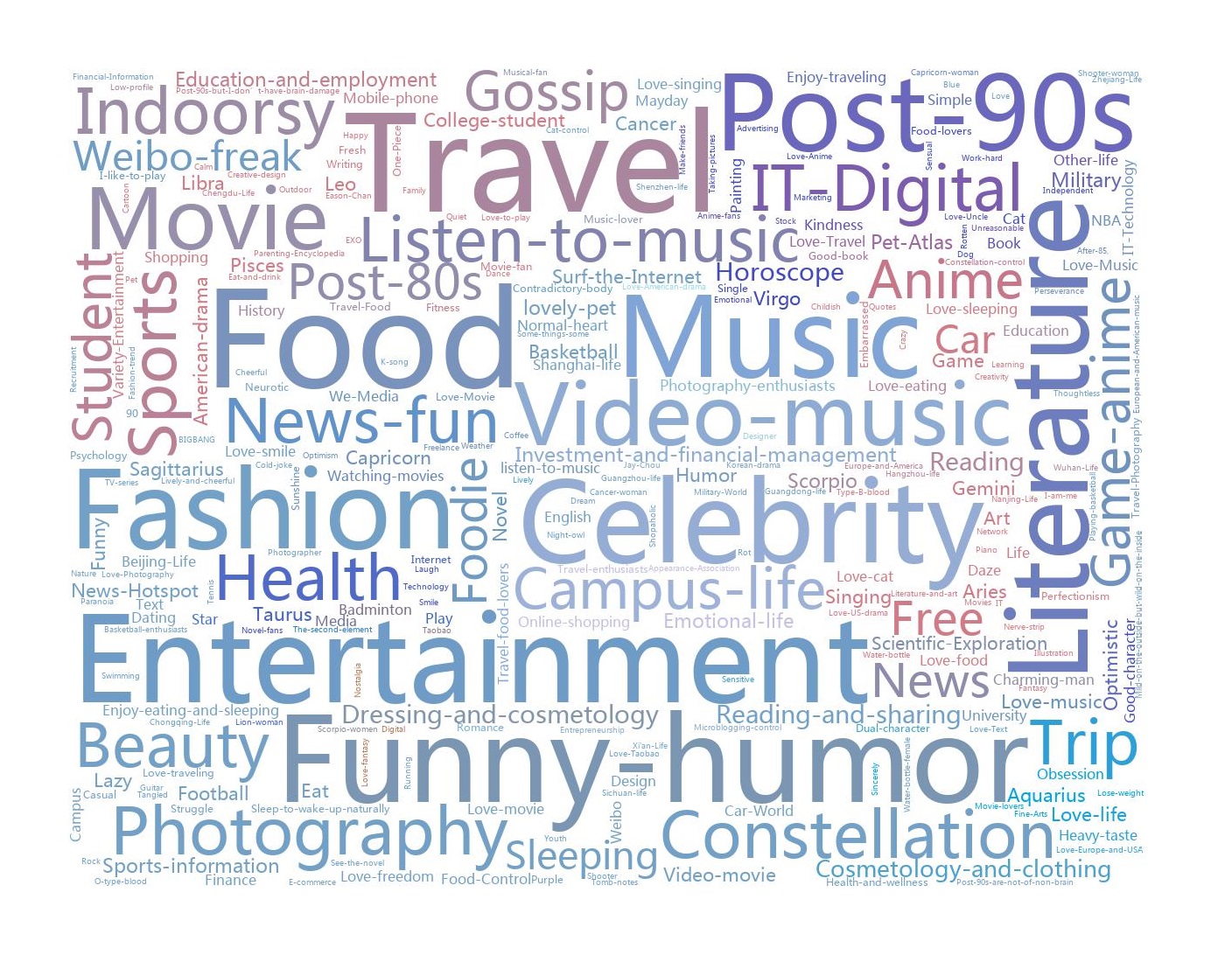} \\
      (a) & (b) & (c)
  \end{tabular}
  \caption{Statistics of personality traits. (a) Distributions of \texttt{Age} and \texttt{Gender} traits. The red and blue bars correspond to female and male speakers, respectively; (b) Distributions of top 21 frequent \texttt{Locations} (provinces); (c) Word cloud visualization of top 250 frequent \texttt{Interest Tags} (translated). The top 10 frequent tags are ``Travel'', ``Food'', ``Entertainment'', ``Funny-humor'', ``Celebrity'', ``Music'', ``Fashion'', ``Literature'', ``Video-music'' and ``Post-90s''.}
  \label{fig:age_gender}
\end{figure*}

Each speaker presented in our dataset has five personality traits: \texttt{Gender}, \texttt{Age}, \texttt{Location}, \texttt{Interest Tags}, and \texttt{Self Description}. Specifically, \texttt{Gender} is a binary-valued trait, i.e., the gender of a speaker can be either ``Male'' or ``Female''; \texttt{Age} is represented by an integer ranging from 8 to 48. Our observation indicates that \texttt{Age} values out of this range are very likely to be ``fake'', i.e., some Weibo users prefer to not to reveal their true ages by providing unreasonable birthdays. Therefore, a speaker with an age out of this range is reserved in our dataset but is given an empty \texttt{Age} value. \texttt{Location} is the province or urban district indicating where the speaker comes from. This trait has 35 different values that cover all the areas of China. \texttt{Interest Tags} is a set of keywords indicating the speaker's hobbies and interests. Each speaker may provides several different tags. In order to reduce the noise of the collected dataset, tags that are shared by less than 10 speakers are ignored in \texttt{PersonalDialog}; \texttt{Self Description} contains some self-provided description utterances of each speaker. It could be his/her quotations or biography; Basic statistics of these personality traits are shown in Table \ref{tab:traits} and Figure \ref{fig:age_gender}.

Note that our data collection process strictly follows the privacy setting of Weibo. All these five personality traits collected in our dataset are publicly available on Weibo. We believe these traits are closely related to speakers' personae in dialogues and speakers contained in our dataset cannot be traced based on the trait information.

\subsubsection{Corpus Size}
In addition to rich personality traits, another appealing feature of \texttt{PersonalDialog} is its large size. Table \ref{tab:dialogue} presents a basic statistic of dialogues in \texttt{PersonalDialog} where there are 20.83M dialogues and 56.25M utterances.

\begin{table}[htb]
\centering
 \caption{Statistics of dialogues in \texttt{PersonalDialog}.}
 \begin{tabular}{ll}
  \toprule
  Total dialogues                       & 20.83 M \\
  Total utterances                      & 56.25 M \\
  Dialogues with more than 4 utterances & 3.43 M  \\
  Average utterances per dialogue       & 2.70    \\
  Average tokens per utterance          & 9.35    \\
  \bottomrule
 \end{tabular}
\label{tab:dialogue}
\vspace{-2mm}
\end{table}


Another advantage of \texttt{PersonalDialog} involves the length of dialogue session. A considerable amount of dialogues (3.43M sessions) in \texttt{PersonalDialog} have multiple turns of conversations. These dialogues can facilitate the research on multi-turn open-domain dialogue systems. To the best of our knowledge, there is still no such publicly available corpus.

\subsubsection{One-to-Many in Dialogue Generation}
Different from machine translation where two sentences from different languages are equivalent in semantic, dialogue generation is essentially a one-to-many mapping problem: for a same post, there are many possible responses dependent on the context, scene, emotional mood, and many other factors. \texttt{PersonalDialog} offers an opportunity to study this challenging research problem since most of existing open-domain dialogue corpora do not contain multiple responses to a post or are of limited scale. Actually, more than 2M posts have at least two replies in our dataset. We believe \texttt{PersonalDialog} will facilitate further studies on developing conversational agents that are able to generate diversified responses.


\subsubsection{Sociolinguistics Phenomena}
\texttt{PersonalDialog} presents a large amount of informal dialogue contents generated in \emph{computer mediated communications} \citep{herring2007faceted}. Together with diversified personality traits, our dataset can facilitate the study of language usages in \emph{computational sociolinguistics} and help to build key components for such research \citep{Nguyen:2016:CSS:3030588.3030595}. In addition, comparing to crowd-sourced corpora, conventional corpora that are collected from social media carry rich social meanings corresponding to each speaker, where the large size of our dataset makes it more feasible to be used in such research. Therefore, \texttt{PersonalDialog} might become a good choice for computational sociolinguistics research.

In fact, in this work, we have explored a preliminary application of \texttt{PersonalDialog} on computational sociolinguistics: the detection of social identities \citep{Nguyen:2016:CSS:3030588.3030595}. Specifically, trait classifiers are devised (introduced in \S\ref{sec:trait_classifier}) to predict the gender, age and location of social media users based on users' Weibo posts. Our classifiers achieve reasonable performance and the corpus facilitates further studies in this direction. In addition, our dataset can also facilitate the modeling of dialectal variations \citep{doyle2014mapping} as well as syntactic and pragmatic variations with respected to \texttt{Age}, \texttt{Gender}, \texttt{Location}, or a mixture of these traits.

It is also worth noting that dialogue datasets used in traditional sociolinguistics researches are usually collected in a way that each speaker explicitly indicates his/her audiences. However, \texttt{PersonalDialog} provides a very different settings because a Weibo post usually do not specify a particular audience, which provides us a chance to validate the findings of prior sociolinguistics studies on new units of analysis \citep{topp2002online}.

\subsection{Data Collection and Filtering} \label{sec:collect}
Our data collection process was separated into two stages to have a smooth initiation and avoid collecting posts from spammers. The first stage collected seed users who commented under some manually chosen Weibo accounts that were specialized at posting news and maintained by dedicated staffs from mass media. The collected seed users were further filtered based on some user statistics such as the number of followers, posts, and followees. About 300k seed users were resulted. The second stage involved collecting Weibo messages posted by these seed users, together with the received comments and personality traits of each commenting user. Note that a tree structure can be constructed based on the \emph{reply-to} relations between these collected comments, and a dialogue session can be obtained by traversing a path from a root comment to each leaf comment. Finally, about 60 million sessions of raw dialogues and 12 million speakers were obtained.

Several pre-processing steps were used to clean these raw dialogues. We first eliminated the dialogues that contained abusive utterances based on a pre-defined abusive word list (containing 3,089 abusive words). A session was discarded if it contained an abusive utterance. Then, all the utterances were tokenized using jieba\footnote{\url{http://github.com/fxsjy/jieba}}. The sessions containing utterances that were too short (less than 3 tokens), too long (more than 40 tokens) or with only stop words were discarded. We also applied some rules to further reduce the noise, such as removing consecutive punctuations and emojis, and truncating dialogues at the utterances that contained only emojis, punctuations, Latin characters, or external links.

\begin{figure}[htbp]
  \centering
  \includegraphics[scale=0.4]{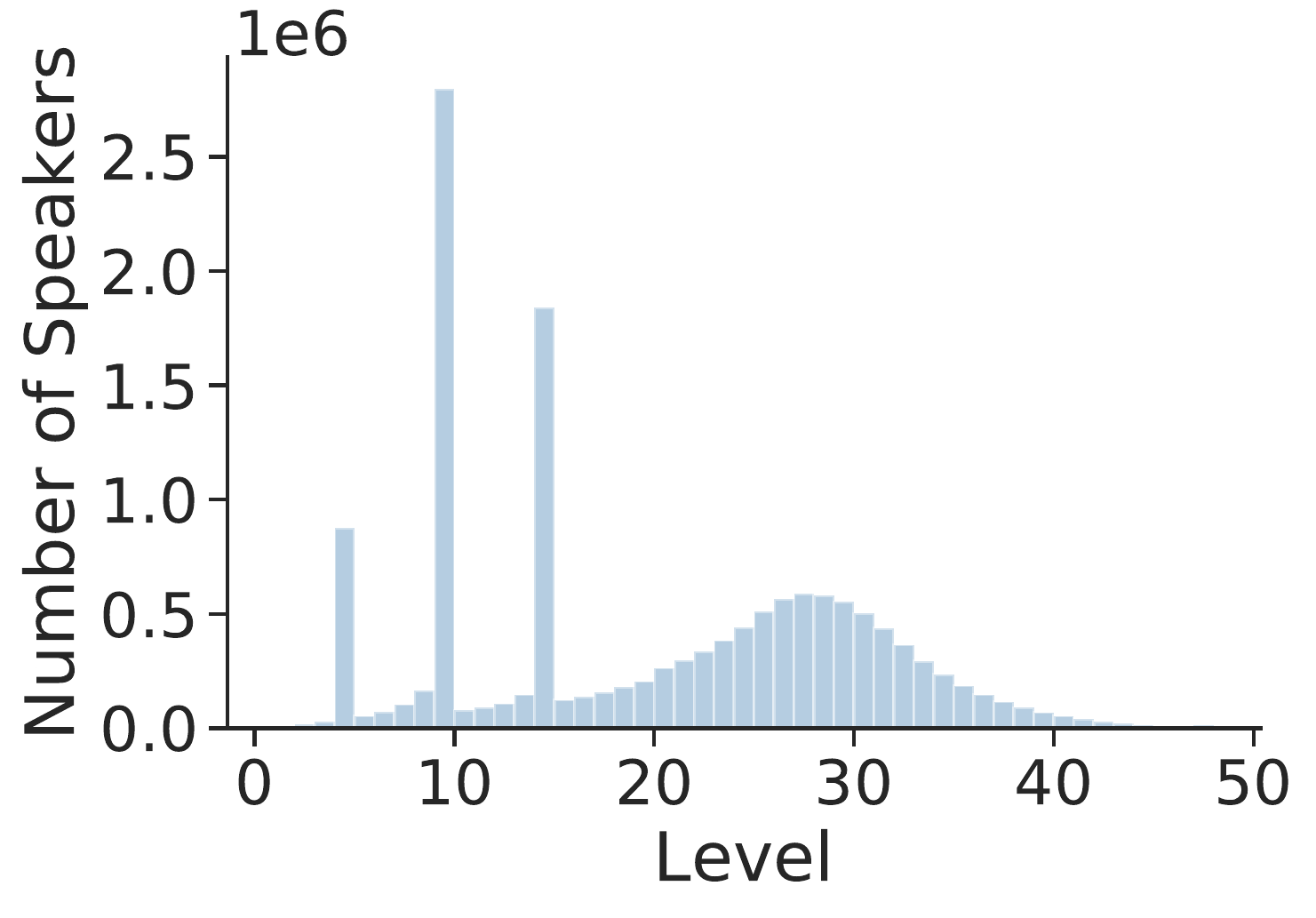}
  \caption{Distribution of the activeness level of collected Weibo users.}
  \label{fig_level}
\vspace{-3mm}
\end{figure}

Another pre-processing step of our data was to filter spammers. In fact, spammer detection is a quite challenging task in social media analysis. It is not our target in this paper to discuss how to accurately detect spammer utilizing the information we have collected. However, what we do care is to ensure the data presented in \texttt{PersonalDialog} are produced by normal human users. Fortunately, the distribution of users' activeness level (shown in Figure \ref{fig_level}) on Weibo sheds a light on this task. The level of a user on Weibo is an indicator of his/her activeness, i.e., a user must be active enough to obtain a high level. \footnote{A detailed explanation for the level system of Weibo can be found here: \url{http://level.account.weibo.com/level/levelexplain}}.
It is interesting to note that there are anomaly peaks at level 4, 9 and 14 in Figure \ref{fig_level}. This may dues to the strict ``upgrade'' rule introduced by Weibo. Specifically, a user has to meet extra requirements (e.g. follows or being followed by a specific number of users) to upgrade in these levels.
We argue that it is hard for most spammers to bypass these levels because it will increase the cost of spamming sharply. We further argue that most spammers are located under level 15, and users with levels higher than 15 are more likely to be regular users. Therefore, dialogues that came from speakers whose activeness level were under 15 were abandoned in \texttt{PersonalDialog}.

\subsection{Personality Trait Classifiers} \label{sec:trait_classifier}
In order to take full advantages of personality traits in personalized dialogue generation, we need to ensure that dialogues collected in \texttt{PersonalDialog} indeed carry trait-related features. 
A natural idea to demonstrate this is to predict the value of personality traits based on dialogues. In particular, to build trait classifiers that take in dialogue texts and predict the value of each trait associated with each speaker. 
The constructed trait classifiers can also be used to evaluate our generation models. Namely, we can determine whether the generated responses reveal certain personality traits using these classifiers.

To this end, three trait classifiers were built, i.e. classifiers for \texttt{Gender}, \texttt{Age}, and \texttt{Location}, respectively. Ideally, these classifiers should be able to identify speakers' \texttt{Gender}, \texttt{Age}, and \texttt{Location} based on the dialogues they issued. In the following sections, we will present details of the data preparation process and classification models.

\subsubsection{Data Preparation}\label{sec:build_corpora}
A naive approach to construct the trait classifier is to predict trait values based on each individual Weibo post (utterance). However, according to a series of crowd-sourcing experiments done by \citet{nguyen2014gender}, speakers may not reveal their persona in each single utterance. Therefore, training trait classifiers using a single utterance as an input generally produces suboptimal performance. 

In order to alleviate above issues, we argue that the value of each trait carried in dialogue texts should be judged based on a set of utterances rather than a single utterance. In this study, we used a concatenation of $n$ utterances as an input to the trait classifier. Specifically, for a given trait, we concatenate every $n$ utterances issued by speakers with a same trait label and use the concatenated texts as inputs to that trait classifier. Obviously, the concatenated texts contain richer information about that trait. This is a commonly used strategy in trait perception tasks performed on social media data \citep{flekova2016analyzing,nguyen2016computational}.

More specifically, taking the gender classifier as an example: since there are only two labels for \texttt{Gender} (``Male'' and ``Female''), we first construct two sets of utterances that are issued by males and females respectively, and then concatenate every $n$ utterances in each set as an input to the gender classifier. We randomly sample 50K such inputs for validation and test, respectively, and use the rest of these inputs for training. Data for other trait classifiers are similarly processed.

Note that the performance of the constructed classifier will be affected by the value of $n$. In fact, the more utterances contained in an input (if $n$ is large), the more evidences will be provided to the classifier to make a decision, which generally leads to higher performance. We have tested different choose of $n$ and found that $n=20$ yields plausible performance in most cases. The accuracy increase will be less than 1\% if $n>20$ for all trait classifiers. Therefore we used $n=20$ for all our classifiers.

\begin{table}[htbp]
\centering
\caption{Statistics of the datasets (with balanced labels) used to build trait classifiers.}
 \begin{tabular}{llllc}
  \toprule
  Trait Type & Train & Validation & Test & Class Num.\\
  \midrule
  \texttt{Age}      & 2.0M & 150K & 150K & 4\\
  \texttt{Gender}   & 5.1M & 65K  & 65K  & 2\\
  \texttt{Location} & 2.9M & 75K  & 75K  & 10\\
  \bottomrule
 \end{tabular}
\label{tab:cls_data}
\end{table}

Moreover, labels used for each classifier deserve further discussion. For \texttt{Gender} classifier, only two labels were used: ``Male'' and ``Female''. Users who did not provide their gender on Weibo were omitted for constructing \texttt{Gender} classifier.
For \texttt{Age} classifier, four labels were used, i.e., we grouped the values of \texttt{Age} into four ranges: ``post-70s'' (born within 1970-1079), ``post-80s''(1980-1989), ``post-90s'' (1990-1999), and ``post-00s'' (born after 2000). This simplification was made because previous studies indicate that it was impractical to predict the exact \texttt{Age} of a speaker with only text from social media~\citep{eckert2017age}. Users with \texttt{Age} values that were out of this range were not used to produce our \texttt{Age} classifier. Similar strategy was used for \texttt{Location} classifier, where ten \texttt{Location} labels were chosen based on the theory of geolinguistics about the dialect area distribution of Chinese \citep{cao2008linguistic}. Instead of predicting exact province, we assigned identical labels for provinces (or districts) corresponding to similar dialect areas.

\begin{table}[!htbp]
\centering
 \caption{Accuracy of trait classifiers.}
 \begin{tabular}{lcccccc}
  \toprule
  Model & \texttt{Gender} & \texttt{Age} & \texttt{Location} \\
  \midrule
  CNN      &       89.71    &      76.84    &     61.02     \\
  LSTM     &       90.23    &      78.02    &     61.69     \\
  RCNN     & \textbf{90.61} &\textbf{78.32} &\textbf{62.04} \\
  \bottomrule
 \end{tabular}
\label{tab:cls_acc}
\vspace{-4mm}
\end{table}

Note that datasets constructed following above settings may suffer from label imbalance issue, i.e. some labels may have remarkably more instances than others. In order to be consistent with the model evaluation scheme presented in \S\ref{sec:auto_eval}, The datasets used to train, validate and test each trait classifier were balanced using the random minority oversampling approach \citep{buda2018systematic}, i.e., instances from minor labels were repeatedly up-sampled. Statistics of the datasets that were used to train each classifier were shown in Table \ref{tab:cls_data}.

\subsubsection{Classification Models}
We trained several classifiers using the constructed datasets, including CNN \citep{Kim2014Convolutional}, LSTM (the states of every time steps of the LSTM layer are averaged before feeding them into a fully connected layer), and RCNN \citep{lai2015recurrent} (the outputs of LSTM states are feed into a CNN layer). Results in Table \ref{tab:cls_acc} show that RCNN achieves the best performance. Therefore, it is used as the classification model in subsequent automatic evaluation of dialogue generation.

The filter sizes used in CNN and RCNN are 2, 3, 4. The feature size for each filter is 128. The hidden size of LSTM and RCNN is 265. Word embeding size is 100 and these models are trained with a dropout rate of 0.8. Note that performances of these classifiers are not sensitive to the choose of hyper-parameters.


%% file: sections/experiment.tex
\section{Experiments} \label{sec:experiments}

\subsection{Data Preparation for Experiments}
To evaluate the influence of personality traits on dialogue generation, we performed single-turn dialogue generation\footnote{Multi-turn dialogue generation can be considered in our framework by encoding additional contexts, and we leave it as the future work.}. To this end, we used 10M sessions of single-turn dialogues (post-response pairs) extracted from \texttt{PersonalDialog}. Three personality traits were considered in our models: \texttt{Gender}, \texttt{Age}, and \texttt{Location}. Similar to the data preparation process presented in \S\ref{sec:build_corpora}, we only considered coarse labels for \texttt{Age} and \texttt{Location}, i.e., 4 labels for \texttt{Age} and 10 labels for \texttt{Location}. We randomly sampled 20,000 dialogue sessions for validation.

To test how well our model can utilize diversified personality traits in different contexts, we constructed four test sets: unbiased set, gender-biased, age-biased, and location-biased set, each of which included 10,000 dialogue sessions. Biased test sets provided us different contexts under which human speakers tend to reveal certain personality trait of themselves. For example, the post-response pair ``Are you a boy or a girl?'' and ``I am a girl'' are \texttt{Gender} biased because most speakers tend to reveal their gender in response to gender-related questions. It will be interesting to see whether our model can learn to incorporate these traits in the generated dialogues under these biased contexts. In this study, dialogues in the unbiased set were randomly sampled. Whereas dialogues in the biased sets were deliberately selected to contain biased responses that carry obvious features related to each trait (like the \texttt{Gender} biased response ``I am a girl'' given in above example). Specifically, the trait label of a biased response should be correctly predicted with high confidence score using associated trait classifier.

The construction process of our biased sets would be straight forward if we had a classifier that toke single response utterances as inputs, i.e., we could feed each individual response to that classifier and selected the correctly predicted responses that had high confidence scores (i.e., maximum value of Softmax outputs) as biased responses. However, as discussed in section \ref{sec:trait_classifier}, our trait classifier toke a concatenation of 20 utterances as an input because not all utterances collected on social-network carry trait-related features. It means that we cannot directly calculate the confidence score for each individual response utterance using our classifier. Moreover, if an input (a concatenation of 20 response utterances) was correctly predicted with high confidence score by our classifier, not all response utterances constituting that input carried trait-related features.

In order to solve above issues, we argue that if a response utterance $r$ is biased, then the input that contains $r$ is more likely to be correctly classified with higher confidence score compared to the input that do not contain $r$. Therefore, we define the confidence score $c(r)$ of each individual response utterance $r$ as the averaged confidence score of all possible inputs that contain $r$. If $c(r)$ is high, it is more likely for the input that contains $r$ to be correctly classified, i.e., for $r$ to be biased.

Apparently, it is unpractical to compute $c(r)$ precisely using its definition. In this study, we obtained an approximation of $c(r)$. Specifically, for a given trait, e.g. \texttt{Gender}, we randomly sampled $N$ post-response pairs $(p_i, r_i), i=1, 2, ..., N$, and for each response sentence $r \in \{r_i, i=1,2,...,N\}$, we constructed $m$ classifier inputs $S_j(r)$, $j=1, 2, ..., m$ containing $r$. Specifically, each input $S_j(r)$ was a concatenation of 20 response sentences, in which $r$ was contained. Assuming $r$ was issued by a female speaker, then the rest 19 response sentences constituting $S_j(r)$ were randomly sampled female-issued responses from $\{r_i, i=1,2,...,N\}$. We fed $S_j(r)$, $j=1, 2, ..., m$ into our \texttt{Gender} classifier and calculated the approximated confidence score for $r$ as:
\begin{equation}
    c'(r) = \frac{1}{m} \sum_{j=1}^{m} \delta(S_j(r)) P(S_j(r)) 
\end{equation}
in which $P(S_j(r)) \in [0,1]$ was the confidence score produced by our \texttt{Gender} classifier when processing $S_j(r)$. $\delta(S_j(r))$ was set to 1 if the label of $S_j(r)$ was correctly predicated, and set to -1 otherwise. In our experiments, we use $N=50,000$ and $m=1,000$. The top 10,000 high scored responses were selected as biased responses and the corresponding post-response pairs were used as the biased set.

We tested each biased set by concatenating every 20 response utterances with a same trait label and feeding these concatenations to our classifier. The classification accuracy of these constructed inputs are reported in the last row of Table \ref{tab:bias-result}. These high accuracy scores indicate that the responses contained in each biased set indeed carry rich trait related features, i.e., can be correctly classified more easily.

Note that we have also tested higher values for $N$ and $m$, which are certainly beneficial for obtaining a more accuracy confidence score. However, these accuracy scores shown in the last row of Table \ref{tab:bias-result} are near-perfect. So we decide to use $N=50,000$ and $m=1,000$ until we can manage to get a better trait classifier.


\subsection{Implementation Details}
We implemented our model and tuned all the hyper-parameters on the validation set. Specifically, the encoder and decoder are 2-layer GRUs with 512 hidden units for each layer. We set the word vocabulary size to 40,000 and the dimension of word vectors to 100. The word vectors are updated during the training process and shared by the encoder and decoder. The embedding size of the persona representation $\vec{v}_p$ is set to 100. The Adam optimizer is used to train our model with a batch size of 120 and a learning rate of 0.001. The training process of each model took about a week on a Titan X GPU machine.

\subsection{Baselines}
We chose several baselines:
\begin{itemize}[leftmargin=1em]
    \item A Seq2Seq model, which does not use any persona features.
    \item Three Group Linguistic Bias Aware (GLBA) models \citep{Wang2017Group}, which respectively incorporate three individual personality traits, namely \texttt{Gender}, \texttt{Age}, and \texttt{Location}.
\end{itemize}
We implemented several variants of our proposed model with different combinations of trait fusion methods and decoding schemes. Three trait fusion methods are attention-based (Att. see \S\ref{sec:trait_att}), average fusion (Avg. see \S\ref{sec:trait_avg}), and concatenation (Concat. see \S\ref{sec:trait_concat}). Two decoding schemes are Persona-Aware Attention (PAA, see \S\ref{sec:paa}) and Persona-Aware Bias (PAB, see \S\ref{sec:pab}). As a result, six variants were tested.

Note that we did not adopt the speaker model introduced by \citet{Li2016_ACL} as our baseline because it requires a large amount of dialogues for each speaker to train a reliable model. Furthermore, we also did not adopt a variation of the speaker model, where speaker embedding is replaced by trait embedding. In fact, the gated bias approach used in GLBA models generally outperforms using trait embedding in the speaker model\citep{jaech2017lowrank_rnn}. 
In other words, GLBA models are stronger baselines for our task.

\begin{table}[!htbp]
\centering
\caption{Automatic evaluation on the unbiased test set with \emph{perplexity} (ppx.), \emph{distinct-1} (dist1), \emph{distinct-2} (dist2) and \emph{trait accuracy} (acc.).}
\begin{tabular}{l<{\centering}p{18pt}<{\centering}p{18pt}<{\centering}p{18pt}<{\centering}p{18pt}<{\centering}p{20pt}<{\centering}p{20pt}<{\centering}}
\toprule
Model                  &      ppx.    & {\small dist1} & {\small dist2}&\texttt{Gender} acc.&\texttt{Age} acc.&\texttt{Loc.} acc. \\ 
\midrule 
Seq2Seq                &     84.07   &  0.0226 & 0.0599  &     50.2      &     25.3    &    10.2    \\  
\texttt{Gender} GLBA   &     79.05   &  0.0287 & 0.0764  &     73.5      &     25.0    &    10.0    \\
\texttt{Age} GLBA      &     79.21   &  0.0285 & 0.0743  &     50.1      &     42.0    &    10.0    \\
\texttt{Location} GLBA &     80.04   &  0.0276 & 0.0689  &     50.1      &     25.1    &    19.6    \\
\midrule                              
Avg. + PAA             &     81.47   &  0.0271 & 0.0746  &     63.5      &     30.2    &    15.4    \\   
Concat. + PAA          &     82.37   &  0.0272 & 0.0735  &     63.4      &     30.6    &    15.8    \\  
Att. + PAA             &     82.26   &  0.0259 & 0.0707  &     70.1      &     29.2    &    14.3     \\  
Avg. + PAB             &     79.46   &  0.0287 & 0.0741  &     76.7      &     37.2    &    20.7     \\   
Concat. + PAB          &     81.51   &  0.0279 & 0.0779  & \textbf{77.9} &     37.5    &    20.8     \\  
Att. + PAB             &\textbf{78.44} & \textbf{0.0293} & \textbf{0.0805}&     77.1      &\textbf{38.9}&\textbf{22.2}\\  
\bottomrule
\end{tabular}
\label{tab:unbias-result}
\vspace{-1mm}
\end{table}

\begin{table}[!htbp]
\centering
\caption{Automatic evaluation on biased test sets with \emph{trait accuracy} (acc.). The \emph{trait accuracy} is obtained using the corresponding biased test set with the originally provided persona traits.}
\label{tab:bias-result}
\begin{threeparttable}
\begin{tabular}{lp{30pt}<{\centering}p{25pt}<{\centering}p{40pt}<{\centering}}
\toprule
Model & \texttt{Gender} acc.& \texttt{Age} acc.& \texttt{Location} acc.\\ 
\midrule
Seq2Seq                &     85.3    &     79.8    &      27.2   \\
\texttt{Gender} GLBA   &     95.5    &     81.6    &      31.8   \\
\texttt{Age} GLBA      &     86.8    &     92.1    &      32.0   \\
\texttt{Location} GLBA &     87.3    &     78.2    &      48.2   \\
\midrule
Avg. + PAA       &     91.1    &     88.5    &      43.3   \\ 
Concat. + PAA    &     91.7    &     88.9    &      44.5   \\ 
Att. + PAA       &     94.0    &     88.3    &      42.5   \\ 
Avg. + PAB       &     94.8    &     91.9    &      48.5   \\ 
Concat. + PAB    &     95.0    &     91.6    &      48.9   \\ 
Att. + PAB       &\textbf{96.0}&\textbf{92.5}&\textbf{50.3}\\ 
\midrule 
Golden responses\tnote{1} &     100.0   &     99.8    &      90.8   \\
\bottomrule
\end{tabular} 
\begin{tablenotes}
\item[1] \small{This score shows the trait accuracy obtained using golden (human-generated) responses in the biased test sets.}
\end{tablenotes}
\end{threeparttable}
\vspace{-4mm}
\end{table}

\subsection{Automatic Evaluation}\label{sec:auto_eval}
We performed automatic evaluation to verify whether our model can incorporate diversified personality traits in dialogue generation.
\subsubsection{Metrics}
\emph{Perplexity} was used to evaluate our model at the content level. Smaller perplexity scores indicate that the model can generate more grammatical and fluent responses. We also used \emph{Distinct}\citep{li2015diversity} to evaluate the diversity of the generated responses. To evaluate how well our models can capture personality traits, we defined \emph{trait accuracy} as the agreement between the expected trait values (i.e., inputs to the personality trait fusion model) and the trait labels predicted by the trait classifiers. A higher \emph{trait accuracy} indicates a stronger ability to incorporate that trait. For example, for the \texttt{Gender} trait, if a set of responses were generated with a ``female'' label, we were expecting these responses can be easily classified to ``female'' by our \texttt{Gender} classifier. Therefore, in order to calculate the \texttt{Gender} accuracy, we first generated responses with different \texttt{Gender} values, and then we followed the process introduced in section \ref{sec:build_corpora} to construct classifier inputs using these generated responses. The classification accuracy of these inputs using our \texttt{Gender} classifier was obtained as the \texttt{Gender} accuracy. Note that in this process, the values of other traits were identical to those of the responder in the test set.


\subsubsection{Results}
Table \ref{tab:unbias-result} shows the performance of each model on the unbiased test set. The \emph{trait accuracy} shown in this table was obtained by assigning the target trait with different values. For example, for the \texttt{Gender} trait, we generated two sets of responses to the same posts with ``Female'' and ``Male'' label, respectively. Moreover, in order to investigate the behaviour of our models in different contexts, we also tested our models on three biased test sets (Table \ref{tab:bias-result}). The \emph{trait accuracy} shown in Table \ref{tab:bias-result} was obtained by assigning the same traits to those of the responder in the test set, i.e., when generating the response for a given post, we provide our models with same traits as the responder in the biased set. We also listed the \emph{trait accuracy} calculated using the responses generated by actual human speakers in the last row of Table \ref{tab:bias-result}. These scores can be used as the upper-bound of the generation models.

Results in these tables show that:
\begin{itemize}[leftmargin=1em]
\item The models equipped with PAB generally outperform the models with PAA on all the metrics. This may be attributed to the fact that PAB can influence the decoding process more directly.

\item GLBA models only perform well on single trait. For example, the \texttt{Gender} GLBA model only achieves good \emph{trait accuracy} regarding to \texttt{Gender}, whereas it degrades remarkably on \texttt{Age} and \texttt{Location} compared to our models. In comparison, our models achieve higher \emph{trait accuracy} with respect to all the traits. This verifies that the trait fusion module is necessary to model diversified traits in different contexts.

\item The model equipped with trait attention (Att.) and PAB obtains the best performance in terms of almost all the metrics, particularly on the biased test sets. This indicates that the trait attention approach facilitates the modeling of diversified traits and it also helps to choose proper traits in different contexts.
\end{itemize}

\subsection{Manual Evaluation}
In order to further evaluate the performance of our models, we performed manual evaluation. Given a post and the personality traits of a responder, we generated responses from all the baseline models and our best performing model (Att. + PAB). These responses were presented to three human annotators along with the specified personality traits and the post.

\begin{table}[!htbp]
\centering
\caption{Manual evaluation with \emph{Fluency} (Flu.) and \emph{Appropriateness} (App.).}
\begin{tabular}{lcc}
\toprule
Model                    &    Flu.      &     App.     \\ 
\midrule                  
Seq2Seq                  &    4.685     &     3.889    \\ 
\texttt{Gender} GLBA     &    4.732     &     3.850    \\
\texttt{Age} GLBA        &    4.792     &     3.898    \\
\texttt{Location} GLBA   &    4.730     &     3.707    \\
\midrule
Att. + PAB               &\textbf{4.822}&\textbf{3.971}\\ 
\bottomrule
\end{tabular}
\label{tab:manual_result}
\vspace{-4mm}
\end{table}

{\small
\begin{table*}[!htbp]
\centering
\caption{Sample responses generated by baselines and our model (Att. + PAB). Words in response are in the same color with the corresponding traits.}
\begin{tabular}{p{50pt}<{\centering}p{45pt}lp{105pt}p{160pt}}
\hline
Post & \shortstack[l]{Traits for Resp.} & Model & Response & Translated Responses \\ 
\hline

\multirow{7}{*}{\shortstack[l]{我的理想男友。\\ Ideal boyfriend \\ for me}}&
\multirow{5}{*}{\shortstack[l]{Post-90s \\ \textcolor{blue}{Male} Haiwai}} 
& Seq2Seq & 给你一个么么哒。 & Give you a kiss \\
&& \texttt{Location} GLBA & 你是我的小男友。 &  You are my little boyfriend \\
&& \texttt{Age} GLBA & 你的男朋友是我的！ & Your boyfriend is mine! \\
&& \texttt{Gender} GLBA & 你是我的\textcolor{blue}{小情人}！ & You are my \textcolor{blue}{little lover}! \\
&& Att. + PAB & 你是我的\textcolor{blue}{优乐美} &  You are my \textcolor{blue}{beloved}\\
\cline{2-5}
&\multirow{2}{*}{\shortstack[l]{Post-90s \\ \textcolor{myorg}{Female} Haiwai}}
& Att. + PAB & 他不只是你的\textcolor{myorg}{理想男友}，呵呵 & He is not just your \textcolor{myorg}{ideal boyfriend}, LOL \\
&& \texttt{Gender} GLBA & 他是\textcolor{myorg}{我的} & He is \textcolor{myorg}{mine} \\
\hline

\multirow{7}{*}{\shortstack[l]{好美的景色。\\The view is so\\ beautiful.}} &
\multirow{5}{*}{\shortstack[l]{\textcolor{blue}{Post-70s} \\ Male Beijing}} 
& Seq2Seq & 谢谢喜欢。快乐。 & Thanks for your liking. Enjoy.\\
&& \texttt{Location} GLBA & 嗯嗯，很美。 & Emm, it is beautiful. \\
&& \texttt{Age} GLBA & 谢谢\textcolor{blue}{好友美评}夸奖！ & Thanks for \textcolor{blue}{my friend's appreciation} and praise! \\
&& \texttt{Gender} GLBA & 是啊，很美！ & Yeah, it is beautiful! \\
&& Att. + PAB & 谢谢\textcolor{blue}{好友美评}，晚上好  & Thanks for \textcolor{blue}{my friend's appreciation}, good evening \\
\cline{2-5}
&\multirow{2}{*}{\shortstack[l]{\textcolor{myorg}{Post-90s} \\ Male Beijing}} & Att. + PAB & 有机会\textcolor{myorg}{来玩呀} & \textcolor{myorg}{Come on}, pay a visit when you have a chance \\
&& \texttt{Age} GLBA & 谢谢\textcolor{myorg}{亲爱的}肯定 & Thank you, \textcolor{myorg}{my dear}\\
\hline

\end{tabular}
\label{tab:example}
\vspace{-2mm}
\end{table*}
}

\subsubsection{Metrics}
Annotators were asked to score a response in terms of two aspects with a five-star scale (1 means not good or not true, while 5 means excellent): 
\begin{enumerate}[leftmargin=2em]
    \item \emph{Fluency}: How do you judge the overall quality of the utterance in terms of its grammatical correctness and fluency?
    \item \emph{Appropriateness}: Do you think the usage of personality traits in the generated response is logical and meet the common practice of a native speaker in daily communications?
\end{enumerate}

\subsubsection{Annotation Statistics}
100 posts were randomly sampled from each of these four test sets (400 posts in total), and 2,000 responses were generated using five models. The inter-rater consistency of the annotation results were measured using the Fleiss' kappa $\kappa$ \citep{randolph2005free}. In particular, the $\kappa$ value for \emph{Fluency} and \emph{Appropriateness} was 0.82 and 0.53, respectively, indicating fairly good agreements between these annotation results.

\subsubsection{Results}
The results are shown in Table \ref{tab:manual_result}. Our model outperforms all our baselines significantly ($t$-test, $p-value$ < 0.05) in both metrics. This indicates that diversified personality traits help to generate more fluent and appropriate responses and our model can learn to incorporate proper personality traits in generated responses.

It is also interesting to see that the Seq2Seq model outperforms some GLBA models in \emph{Appropriateness}. We argue that these GLBA models try to emphasize single personality trait in each utterance they generate, resulting in sub-optimal performance in producing logical and appropriate responses. In fact, different traits should be embodied in different contexts, and sometimes we do not need to address any trait in the response.

\subsection{Case Study}
Some sampled cases are shown in Table~\ref{tab:example}. Words in responses are in the same color with the corresponding traits. It can be seen that our model can generate responses incorporating certain traits and can choose proper personality traits for different contexts, whereas the Seq2seq model tends to generate universal responses and the GLBA model can only consider a single trait. For the first post, both our model and the \texttt{Gender} GLBA model incorporate proper trait (i.e., \texttt{Gender}) in the generated responses, i.e., these models can act as either ``Male'' (colored in blue) or ``Female'' (colored in orange) when generating responses, while other models can only generate responses with random traits or universal responses. Furthermore, responses to the second post are associated with the \texttt{Age} trait, which is usually expressed in a more implicit way. Our model and the \texttt{Age} GLBA model can incorporate stylistic features related to \texttt{Age}. Specifically, an elder agent (``Post-70s'', colored in blue) tends to use rigorous and formal expressions whereas a younger agent (``Post-90s'', colored in orange) uses casual and informal phrases.

\begin{figure}[!htbp]
  \centering
\begin{tabular}{c}
      \includegraphics[scale=0.34]{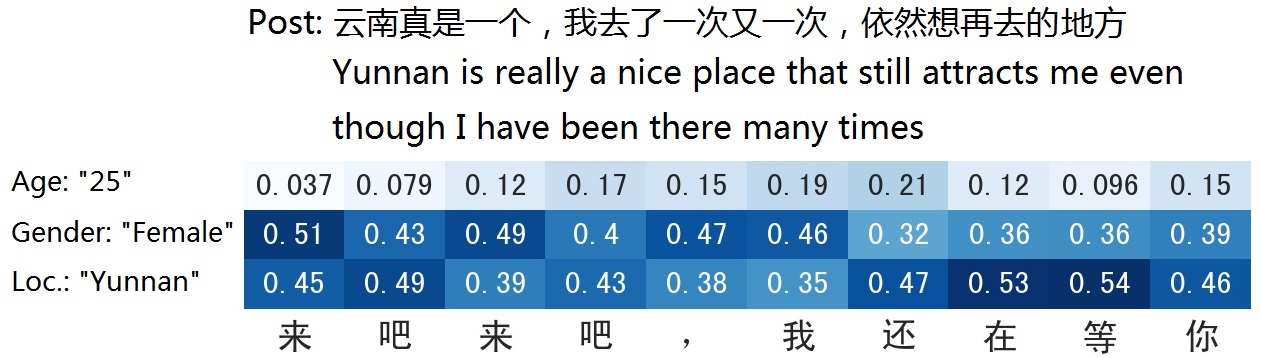} 
  \end{tabular}
  \caption{Visualization of trait attention scores for our model (Att. + PAB). The generated response is ``来吧来吧，我还在等你'' (\emph{Come on come on, I am still waiting for you (in Yunnan)}). ``云南''(``Yunnan'') is a province in China.}
  \label{fig:trait_att}
\vspace{-3mm}
\end{figure}

The visualization of trait attention scores further proves that the trait fusion module helps to model diversified personality traits. As shown in Figure~\ref{fig:trait_att}, when decoding the first four words (i.e., ``come on come on''), which are commonly used by {\it females}, the trait attention scores regarding to \texttt{Gender} are higher. When generating contents related to locations (i.e. ``still waiting for you (in Yunnan)''), the trait attention scores regarding to \texttt{Location} are higher.

%% file: sections/conclusion.tex
\section{Privacy Protection}
Privacy is an important issue for most datasets that are collected from social media, particularly when personal information is involved. In order to protect the privacy of each speaker in our dataset, we designed several anonymization schemes following a critical principle that {\it the speaker's identity should not be traceable}:

\noindent(1) The IDs for speakers and Weibo posts were masked;

\noindent(2) The dialogues that involve explicit references to other users (in particular, using ``@'') were abandoned;

\noindent(3) A de-lexicalization operation was performed by replacing all the numbers with a $\langle {\rm NUM} \rangle$ placeholder. It helped to hide more details of each speaker, such as phone numbers or addresses.

The above proposed schemes can effectively anonymize the private information related to each speaker. Moreover, in order to further protect speakers' privacies, we limit the use of \texttt{PersonalDialog} to be strictly constrained to academic researchers without any attempts to de-anonymize the released data. Anyone who want to use the data has to sign a contract to obey these rules strictly.

\section{Conclusion and Future Work}
In this paper, we investigate a novel task to generate personalized dialogue responses by considering explicitly represented personality traits. To facilitate such research, we first construct a large-scale (20.83M sessions) multi-turn dialogue dataset, \texttt{PersonalDialog}, from real social conversations. The dataset contains various traits (e.g. \texttt{Age}, \texttt{Gender}, \texttt{Location} and \texttt{Interest Tags}) of a large number of speakers (8.47M speakers). Then, we present personalized dialogue generation models to capture and address personality traits in the dialogue generation process. These models apply a trait fusion module to obtain the persona representation of a speaker, and two approaches to address persona-related features in the decoding process: namely persona-aware attention mechanism which dynamically generates context vectors conditioned on the persona representation, and persona-aware bias which manipulates the final generation distribution directly. Automatic and manual evaluation shows that our models can incorporate richer traits in dialogue generation and can learn to choose proper traits in different contexts.

We demonstrate simple models for personalized dialogue generation, and they can serve as baselines for further studies in this research direction since the topic is still in its infancy. The corpus \texttt{PersonalDialog} will facilitate not only the study of personalized dialogue systems, but also other research areas such as sociolinguistics or social science.